\pdfoutput=1

\documentclass[11pt]{article}

\usepackage[nohyperref]{EMNLP2023}

\usepackage{times}
\usepackage{latexsym}
\usepackage{float}
\usepackage{array} 
\usepackage{booktabs}
\usepackage{graphicx}
\usepackage{rotating}
\usepackage{multirow}
\usepackage{caption}

\usepackage[T1]{fontenc}

\usepackage[utf8]{inputenc}

\usepackage{microtype}

\let\ogendabstract\endabstract
\usepackage{arabtex}
\usepackage{utf8}

\let\endabstract\ogendabstract

\usepackage{inconsolata}
\usepackage[noapacite,notextfont,nolineno]{mnn_style}

\title{
Rosetta Stone at the Arabic Reverse Dictionary Shared Task: \\
A Hop From Language Modeling To Word--Definition Alignment
}

\author{
    Ahmed ElBakry\textsuperscript{*} \\
    Microsoft \\
    \texttt{ahmedelbakry@microsoft.com} \\\And
    Mohamed Gabr \\
    Microsoft \\
    \texttt{mohamed.gabr@microsoft.com} \\\AND
    Muhammad ElNokrashy \\
    Microsoft \\
    \texttt{muhammad.elnokrashy@microsoft.com} \\\And
    Badr AlKhamissi\textsuperscript{*} \\
    EPFL \\
    \texttt{badr.alkhamissi@epfl.ch}
}

\begin{document}
\maketitle
\def\thefootnote{*}\footnotetext{Equal contribution}\def\thefootnote{\arabic{footnote}}
\begin{abstract}

A Reverse Dictionary is a tool enabling users to discover a word based on its provided definition, meaning, or description. Such a technique proves valuable in various scenarios, aiding language learners who possess a description of a word without its identity, and benefiting writers seeking precise terminology. These scenarios often encapsulate what is referred to as the "Tip-of-the-Tongue" (TOT) phenomena. In this work, we present our winning solution for the Arabic Reverse Dictionary shared task. This task focuses on deriving a vector representation of an Arabic word from its accompanying description. The shared task encompasses two distinct subtasks: the first involves an Arabic definition as input, while the second employs an English definition. For the first subtask, our approach relies on an ensemble of finetuned Arabic BERT-based models, predicting the word embedding for a given definition. The final representation is obtained through averaging the output embeddings from each model within the ensemble. In contrast, the most effective solution for the second subtask involves translating the English test definitions into Arabic and applying them to the finetuned models originally trained for the first subtask. This straightforward method achieves the highest score across both subtasks.\footnote{\url{https://github.com/bkhmsi/RashidRevDict}}

\end{abstract}

\section{Introduction}

The Tip-of-the-Tongue phenomena, as explained by the authors of \citet{BROWN1966325}, is ``a state in which one cannot quite recall a familiar word but can recall words of similar form and meaning''. A straightforward way to solve this problem, is to have a reverse dictionary; a system that takes a description provided by the user as an input, and outputs the word \citep{dictionaryreversedef}. 

The initial solutions were heuristic-based. In their work, \citet{heursiticrevdict1} suggested a method where the tokens in the user-provided description are compared to all dictionary definitions. The system then returns the word with the highest token match. Their method implements different retrieval efficiency tweaks to overcome the issue of excessive time complexity resulting from the comparison operation. 


Recent approaches employ neural-based models, since that are capable of better capturing the semantics of an input description, in contrast to the earlier solutions mentioned, which relied on word overlap. In their work, \citet{RNNRevDict} suggest utilizing a recurrent-neural-network (RNN) to generate a vector representation based on the provided definition. This representation is then compared against a set of word embeddings to select the closest word to return to the user.

The issue of low-frequency words is one of the main challenges of building a reverse dictionary, since these words are the ones that are the less trained and thus have a worse representation compared to more frequent words. \citet{lowfreqimprovedRevDict} tackle this problem by handcrafting predictors that extract features inspired by the thought process undergone by humans to get a word given its description. 
 
Polysemy, which is the coexistence of many possible meanings for a word, is another obstacle when building reverse dictionaries. Most of the previously mentioned solutions rely on a a set of static word representation builders such as Word2Vec, which hinders the accuracy of such models. This motivates the use of pretrained language models to produce embeddings that vary based on context. The authors of \citet{bertforrevdict} probed BERT \cite{devlin-etal-2019-bert} to predict the word representation, alleviating the issue of polysemy.

\begin{table*}[h]
\centering

\begin{tabular}{lcc}
\toprule
& \textbf{Example Word} & \textbf{Example Definition} \\
\midrule
Task 1 & {\RL{t.hnnana `lyh}} & {\RL{tar.ham, ta`.taf `lyh w ra.hmah}} \\ 
Task 2 & {\RL{zwwar alkalaama}} & To knowingly and willfully make a false statement of witness while in court \\ 

\bottomrule
\end{tabular}
\caption{Word-Definition Pairs Illustrating Subtasks 1 and 2.}
\label{tab:examples}
\end{table*}

Reverse dictionaries can also be cross-lingual; where one aims to retrieve a word in language \texttt{X} based on a description provided in language \texttt{Y}. Employing any of the previously mentioned solutions for a multilingual context necessitates the alignment of word vectors across different languages, a challenging task even for two languages, not to mention when dealing with multiple languages. The authors of \citet{bilingualrevdict} built a collection of bilingual reverse dictionaries using Wiktionary. Other solutions used existing multilingual models, such as mBERT, to reduce the issue of cross-lingual alignment \citet{bertforrevdict}.

The shared task of the Arabic Reverse Dictionary provides a set of words, along with their \code{SGNS} \cite{word2vecsgns} and \code{ELECTRA} \citet{electra} vector representations, and their corresponding definition, in both Arabic and English. A set of Arabic-English word mappings is also supplied to help in building an alignment scheme. The goal of subtask 1 is to predict the \code{SGNS} and \code{ELECTRA} embeddings of the set of Arabic words, given the input Arabic definition. Subtask 2 has the same goal except that an English definition is provided instead of Arabic.

The shared task setup poses multiple obstacles that our solutions attempt to overcome: (1) the small size of the set of aligned words, (2) the blackbox nature of the \code{SGNS} and \code{ELECTRA} word embedding generation pipeline.

Our solution simply finetunes multiple Arabic BERT-based pretrained models to predict an embedding for each word. 



\section{Datasets}

The provided data can be categorised into three distinct datasets.

\begin{enumerate}
    \item The Arabic Language Dictionary is a dataset with $58,010$ entries, where each of datapoint contains a word, an \code{ELECTRA} embedding, an \code{SGNS} embedding, a \code{gloss} (definition of the word), a \code{POS} tag, an \code{ID} and an \code{English ID} where applicable to link with the alignment data.
    \item The English Language Dictionary dataset has $63,596$ datapoints, with the same columns as the Arabic Dictionary except that the embeddings are obtained from English words and not Arabic.
    \item The English Arabic Mapped Dictionary has $4,355$ datapoints in total. Each point has the Arabic and English glosses, Arabic and English IDs, Arabic and English words, and the Arabic embeddings.
\end{enumerate}

The first and third datasets are split into \code{train}, \code{dev} and \code{test} sets by the organizers. The English language dictionary however, isn't provided with such divisions. Therefore, we manually split the English dictionary ourselves. \Tableref{table-data-sizes} shows the split sizes of each dataset. The English dictionary was divided into two sets only, \code{train} and \code{dev}, since there was no need for a test set in our case, and no submission to be made with this dictionary. 

\begin{table}[H]
\centering
\begin{tabular}{lccc}
\toprule
 & \textbf{Train} & \textbf{Dev} & \textbf{Test} \\ 
\midrule
\textbf{Ar Dict} & 45,200 & 6,400 & 6,410 \\ 
\textbf{Ar-En Map} & 2,843 & 299 & 1,213 \\
\textbf{Ar Dict} & 50,877 & 12,719 & N/A \\ 
\bottomrule
\end{tabular}
\caption{Statistics about Data Sizes}
\label{table-data-sizes}
\end{table}

\section{System}






\subsection{Subtask 1: Arabic Definitions to Arabic Embeddings}

In this subtask, we finetune four Arabic BERT-based pretrained models. Namely: (1) \modelname{MARBERTv2} \citep{abdul-mageed-etal-2021-arbert}, (2) \modelname{AraBERTv2} \cite{antoun2020arabert}, (3) \modelname{CamelBERT-MSA} and (4) \modelname{CamelBERT-Mix} \cite{inoue-etal-2021-interplay}. Each model is finetuned twice for this subtask, once for predicting the corresponding \code{SGNS} embedding for each input definition, and the other time for predicting the corresponding \code{ELECTRA} embedding. The final representation is computed by taking the embedding of the \texttt{CLS} and passing it through a two-layer dense network with a \code{Tanh} activation function in between. The model is trained by optimizing the Mean Squared Error (MSE) between the ground-truth representation and the predicted one. For the learning rate scheduling policy, we used OneCycleLR  \citep{DBLP:journals/corr/abs-1708-07120}. Throughout the finetuning process, we evaluate on the devset after every epoch, and take the checkpoint with the highest cosine similarity score. Table \ref{table:hyperparameters} shows the values of the hyperparameters used during finetuning.  


To identify the optimal ensemble of our fine-tuned models, we select the model combination that exhibited the highest performance on the devset, determined by the cosine similarity metric, as our final solution. Tables \ref{tab:task1_comb} and \ref{tab:task2_comb} in the Appendix shows the performance of all model combinations on the devset. The final representation of each ensemble is taken by averaging the predicted embedding of each for a given input definition.

\begin{table}[H]
\centering
\begin{tabular}{cc} 
 \toprule
 \textbf{Hyperparameter} & \textbf{Value} \\ 
 \midrule
 Batch Size & 100  \\ 
 lr & 1.0e-4  \\ 
 Learning Rate Sched. & OneCycleLR  \\ 
 $pct$ & 0.2  \\ 
 $f_{initial}$ & 25  \\ 
 $f_{final}$ & 100  \\ 
 Weight Decay & 1.0e-4  \\ 
 Epochs & 20  \\ 
 Optimizer & AdamW  \\ 
 \bottomrule
\end{tabular}
\caption{Hyperparameters Used}
\label{table:hyperparameters}
\end{table}





\subsection{Subtask 2: English Definitions to Arabic Embeddings}
\label{sec:solutiontask2}

Subtask 2 differs from subtask 1 by utilizing an English definition as input instead of Arabic, with the objective of generating the embedding representation of the Arabic word as output. Several approaches were explored in pursuit of optimizing the system for superior output embedding quality.

\begin{figure}[h]
    \centering
    \includegraphics[width=0.60\linewidth]{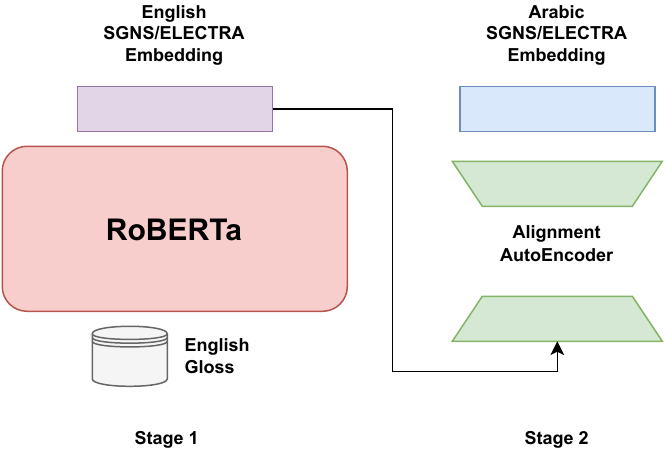}
    \caption{Method Explored for Subtask 2 (Section \ref{sec:solutiontask2})}
    \label{fig:auto-encoder}
\end{figure}

\paragraph{Cross-Lingual Alignment}
This method involves a two-step learning process. First, we leverage the English Language Dictionary to learn to generate the English embeddings from their corresponding English definition. Then the second stage utilize the English Arabic Mapped Dictionary to learn an alignment function between both language representations. Figure \ref{fig:auto-encoder} shows an illustration of this model. The motivation behind this is that the English pretrained models often yield superior representations compared to their Arabic counterparts due to their training on larger corpora. Here, we used \modelname{RoBERTa} \citep{roberta} to obtain English embeddings, following the same procedure as in subtask 1, and then utilizing an autoencoder model to transform these embeddings into their Arabic representations. Both the encoder and the decoder of the Autoencoder consist of two linear layers with ReLU in between. The input and output dim is 256 and the hidden dim is 32. However, the efficacy of converting an English representation into an Arabic one is contingent upon the quantity of aligned data points available in the provided resources.


\paragraph{Translate-Test} 
Our solution for subtask 2 that yielded the best results was inspired from \cite{Artetxe2023RevisitingMT}. In their work, they show that machine translating a non-English test sets into English and then running inference on a monolingual English model can exhibit superior performance compared to using a multilingual model, such as \modelname{XLM-R} \cite{conneau-etal-2020-unsupervised}, on the original data zeroshot. Similarly, we use the Arabic translation of the English definitions as input to our finetuned Arabic models. This approach enables the reuse of models and solutions that were initially developed for subtask 1
.

\section{Results}
\begin{table*}[h]
\centering
\begin{tabular}{ccccccc}
\toprule
\textbf{Subtask} & \textbf{Embedding} & \textbf{MSE} & \textbf{Cosine} & \textbf{Rank} & \textbf{P@1} & \textbf{P@10} \\
\midrule 
\multirow{2}{*}{Subtask 1} & Electra & 0.152 / 0.161 & 0.645 / 0.637 & 0.242 / 0.214 & 0.031 / 0.034 & 0.099 / 0.114 \\
                           & SGNS & 0.030 / 0.035 & 0.605 / 0.552 & 0.254 / 0.281 & 0.445 / 0.414 & 0.597 / 0.540 \\
\midrule 
\multirow{2}{*}{Subtask 2} & Electra & 0.170 / 0.180 & 0.659 / 0.624 & 0.127 / 0.204 & 0.185 / 0.120 & 0.407 / 0.355 \\
                           & SGNS & 0.053 / 0.048 & 0.400 / 0.387 & 0.320 / 0.372 & 0.312 / 0.316  & 0.375 / 0.389 \\
\bottomrule
\end{tabular}
\caption{Results on TestSet / DevSet for Both Subtasks. \textbf{MSE} is Mean-Squared-Error. \textbf{P} is Precision.}
\label{table:test_results}
\end{table*}



Table \ref{table:test_results} displays the results obtained on the test set across all metrics reported in the shared task. Interestingly, the best ensemble on both subtasks was done by taking the average of the \modelname{CamelBERT-MSA} and \modelname{MARBERTv2} output embeddings.

\subsection{Subtask 1}

Table \ref{tab:task1_comb} shows the results on the devset that we can use for further analysis. It clearly illustrates that ensembles, regardless of the combination, enhance the scores in comparison to using individual models. Furthermore, it is evident that results involving \modelname{CamelBERT-Mix} tend to be less favorable than those involving \modelname{CamelBERT-MSA}. This observation aligns with the dataset's nature, which predominantly features MSA definitions, thus minimizing dialectal content.

Through examining the scores of ensembles and systems incorporating \modelname{MARBERTv2} compared to those that do not, we can conclude that \modelname{MARBERTv2} stands out as the most effective model to employ or include in an ensemble among all the tested Arabic pretrained transformers.

\subsection{Subtask 2}

The findings from Subtask 1 are applicable to Subtask 2, and this consistency can be attributed to the reuse of models initially developed in Subtask 1 for Subtask 2.


\section{Discussion}

\paragraph{Exploring Cross-Lingual Alignment Further}
In the pursuit of optimizing our approach for the Arabic Reverse Dictionary shared task, we implemented a cross-lingual alignment method, as detailed in section \ref{sec:solutiontask2}. This method allowed us to bridge the gap between English and Arabic definitions, by leveraging the aligned dictionary provided as part of the shared task. Further exploration and refinement could yield promising results in that direction.

\paragraph{Augmenting Training Data Through Self-Synthesis}
In another set of experiments, we explored a very different approach that requires further investigation in future work. The idea is to finetune of an encoder-decoder model, such as AraT5 \cite{nagoudi-etal-2022-arat5}, jointly on two interconnected tasks. The first task involves predicting the word embeddings from the encoder side, while the second task entails predicting the corresponding definition on the decoder side based on an the input word. This approach presents an intriguing opportunity to generate diverse definition-embedding pairs using a single model, which could subsequently be harnessed for more robust finetuning. This self-synthesis approach could potentially lead to better system performance by expanding the training set. 

\section{Conclusion}





In this paper, we present our winning solution to the Arabic Reverse Dictionary shared task. The objective is to derive an Arabic word representation based on a provided definition, which can be in either Arabic or English.

Our approach simply leverages several language models pretrained on Arabic datasets. Through finetuning and ensembling the trained models, our method is capable of capturing the underlying semantics of the input definitions as well as correcting small errors done by single models. 

For the first subtask, we achieve the best results by fine-tuning four Arabic pretrained language models twice, one for predicting the \code{Electra} embedding and once for the \code{SGNS} one. This involves minimizing the discrepancy between the predicted embedding and the model's final representation using an MSE loss function.

In the second subtask, our most effective solution is to repurpose the models initially developed for the first subtask by translating the English test set definitions into Arabic.

\section*{Limitations}

One notable limitation is related to the second subtask, where our approach involves translating English definitions to Arabic. The results of this paper used the existing Arabic translations that comes English test set. Therefore, we have not investigated the quality of machine translation models, which can significantly influence the system's effectiveness, as inaccuracies or nuances lost in translation may affect results. Moreover, the generalization of our models to broader or different distributions may be constrained, as they are optimized on specific datasets. To achieve wider applicability, we might necessitate further finetuning on more diverse data sources. Furthermore, our choice of evaluation metrics can influence the perceived performance of the system, and different metrics may reveal varying aspects of its utility in practical applications. It is essential to consider these limitations when assessing the robustness and adaptability of our approach.


\bibliography{custom}
\bibliographystyle{acl_natbib}

\appendix

\section{Cross-Lingual Alignment Model}

\section{Results on Devset}
Please refer to next page.

\begin{table*}[h!]
\centering
\begin{tabular}{lllllll}
\toprule
\multirow{2}{*}{\textbf{Models}} & \multicolumn{3}{c}{\textbf{Electra}} & \multicolumn{3}{c}{\textbf{SGNS}} \\
& MSE & Cosine & Rank & MSE & Cosine & Rank \\
\midrule
arabert & 0.1695 & 0.6124 & 0.2491 & 0.0368 & 0.4853 & 0.3321 \\
camelbert-mix & 0.1689 & 0.6134 & 0.2667 & 0.0389 & 0.4911 & 0.3221 \\
camelbert-msa & 0.1681 & 0.6166 & 0.2421 & 0.0379 & 0.4947 & 0.3213 \\
marbert & 0.1661 & 0.6265 & 0.2108 & 0.0370 & 0.5485 & 0.2782 \\
arabert,camelbert-mix & 0.1656 & 0.6228 & 0.2525 & 0.0360 & 0.5045 & 0.3237 \\
arabert,camelbert-msa & 0.1650 & 0.6247 & 0.2400 & 0.0358 & 0.5052 & 0.3235 \\
arabert,marbert & 0.1618 & 0.6355 & 0.2175 & 0.0337 & 0.5511 & 0.2836 \\
camelbert-mix,camelbert-msa & 0.1653 & 0.6239 & 0.2496 & 0.0370 & 0.5036 & 0.3208 \\
camelbert-mix,marbert & 0.1622 & 0.6341 & 0.2267 & 0.0348 & 0.5502 & 0.2817 \\
\textbf{camelbert-msa,marbert} & 0.1614 & 0.6365 & 0.2144 & 0.0345 & 0.5519 & 0.2812 \\
arabert,camelbert-mix,camelbert-msa & 0.1642 & 0.6272 & 0.2455 & 0.0357 & 0.5095 & 0.3221 \\
arabert,camelbert-mix,marbert & 0.1616 & 0.6356 & 0.2286 & 0.0339 & 0.5466 & 0.2862 \\
arabert,camelbert-msa,marbert & 0.1610 & 0.6371 & 0.2204 & 0.0338 & 0.5472 & 0.2860 \\
camelbert-mix,camelbert-msa,marbert & 0.1614 & 0.6361 & 0.2268 & 0.0346 & 0.5452 & 0.2849 \\
arabert,camelbert-mix,camelbert-msa,marbert & 0.1613 & 0.6363 & 0.2287 & 0.0341 & 0.5421 & 0.2895 \\
\bottomrule
\end{tabular}

\caption{Performance Analysis on the Devset of Subtask-1 Using Various Model Ensembles.}
\label{tab:task1_comb}
\end{table*}

\begin{table*}[h!]
\centering
\begin{tabular}{lllllll}
\toprule
\multirow{2}{*}{\textbf{Models}} & \multicolumn{3}{c}{\textbf{Electra}} & \multicolumn{3}{c}{\textbf{SGNS}} \\
& MSE & Cosine & Rank & MSE & Cosine & Rank \\
\midrule
arabert & 0.1879 & 0.6014 & 0.2369 & 0.0491 & 0.3500 & 0.3925 \\
camelbert-mix & 0.1894 & 0.5974 & 0.2482 & 0.0520 & 0.3504 & 0.3956 \\
camelbert-msa & 0.1860 & 0.6066 & 0.2167 & 0.0498 & 0.3580 & 0.3845 \\
marbert & 0.1858 & 0.6108 & 0.2115 & 0.0530 & 0.3818 & 0.3739 \\
arabert,camelbert-mix & 0.1848 & 0.6104 & 0.2354 & 0.0486 & 0.3619 & 0.3920 \\
arabert,camelbert-msa & 0.1829 & 0.6149 & 0.2218 & 0.0479 & 0.3640 & 0.3855 \\
arabert,marbert & 0.1806 & 0.6226 & 0.2106 & 0.0478 & 0.3867 & 0.3752 \\
camelbert-mix,camelbert-msa & 0.1842 & 0.6117 & 0.2245 & 0.0494 & 0.3619 & 0.3894 \\
camelbert-mix,marbert & 0.1821 & 0.6186 & 0.2175 & 0.0493 & 0.3838 & 0.3776 \\
\textbf{camelbert-msa,marbert} & 0.1800 & 0.6238 & 0.2038 & 0.0484 & 0.3874 & 0.3715 \\
arabert,camelbert-mix,camelbert-msa & 0.1827 & 0.6160 & 0.2248 & 0.0481 & 0.3662 & 0.3890 \\
arabert,camelbert-mix,marbert & 0.1808 & 0.6222 & 0.2162 & 0.0476 & 0.3841 & 0.3778 \\
arabert,camelbert-msa,marbert & 0.1794 & 0.6255 & 0.2075 & 0.0472 & 0.3860 & 0.3736 \\
camelbert-mix,camelbert-msa,marbert & 0.1805 & 0.6228 & 0.2135 & 0.0482 & 0.3836 & 0.3761 \\
arabert,camelbert-mix,camelbert-msa,marbert & 0.1800 & 0.6240 & 0.2125 & 0.0474 & 0.3830 & 0.3782 \\
\bottomrule
\end{tabular}

\caption{Performance Analysis on the Devset of Subtask-2 Using Various Model Ensembles.}
\label{tab:task2_comb}
\end{table*}

\end{document}